\documentclass[review]{elsarticle}

\usepackage[OT4]{fontenc}

\usepackage[utf8]{inputenc} %unicode support

\usepackage{lineno,hyperref}
\usepackage{algorithm}
\usepackage{graphicx}
\usepackage[noend]{algpseudocode}
\usepackage{amsthm,amsmath}
\usepackage{amssymb}
\usepackage{array}
\usepackage{color}
\usepackage{enumerate}
\usepackage{caption}
\usepackage{adjustbox}
\usepackage{setspace}
\usepackage[section]{placeins}
\usepackage{gensymb}
\usepackage{mathtools}
\usepackage{amsthm}
\usepackage{lineno}
\usepackage{rotating}

\usepackage{tabularx}
\newcolumntype{L}[1]{>{\raggedright\arraybackslash}p{#1}}
\newcolumntype{C}[1]{>{\centering\arraybackslash}p{#1}}
\newcolumntype{R}[1]{>{\raggedleft\arraybackslash}p{#1}}

\usepackage{multirow}

\newcommand{\mincov}{\textit{mincov}}

\newcommand{\IF}{\textbf{IF }}
\newcommand{\THEN}{\textbf{ THEN }}
\newcommand{\AND}{\ \pmb{\wedge}\ }

\newcolumntype{R}[2]{%
	>{\adjustbox{angle=#1,lap=\width-(#2)}\bgroup}%
	l%
	<{\egroup}%
}
\begin{document}

\begin{frontmatter}

\title{SCARI: Separate and Conquer Algorithm for Action Rules and Recommendations Induction}
%\tnotetext[mytitlenote]{Fully documented templates are available in the elsarticle package on \href{http://www.ctan.org/tex-archive/macros/latex/contrib/elsarticle}{CTAN}.}

%% Group authors per affiliation:
%% or include affiliations in footnotes:
%% or include affiliations in footnotes:
\author[polsl]{Marek Sikora\corref{mycorrespondingauthor}}
\ead{marek.sikora@polsl.pl}
%\ead[url]{www.elsevier.com}

\author[polsl]{Pawe\l{} Matyszok}
\ead{pawel.matyszok@polsl.pl}

\author[polsl]{\L{}ukasz Wr\'obel}
\ead{lukasz.wrobel@polsl.pl}

\cortext[mycorrespondingauthor]{Corresponding author}
\address[polsl]{Faculty of Automatic Control, Electronics and Computer Science, Silesian University of Technology, Akademicka 16, 44-100 Gliwice, Poland}
%\address[emag]{Institute of Innovative Technologies, EMAG, Leopolda 31, 40-189 Katowice, Poland}

%\author{Elsevier\fnref{myfootnote}}
%\address{Radarweg 29, Amsterdam}
%\fntext[myfootnote]{Since 1880.}

%% or include affiliations in footnotes:
%\author[mymainaddress,mysecondaryaddress]{Elsevier Inc}
%\ead[url]{www.elsevier.com}

%\author[mysecondaryaddress]{Global Customer Service\corref{mycorrespondingauthor}}
%\cortext[mycorrespondingauthor]{Corresponding author}
%\ead{support@elsevier.com}

%\address[mymainaddress]{1600 John F Kennedy Boulevard, Philadelphia}
%\address[mysecondaryaddress]{360 Park Avenue South, New York}

\begin{abstract}
This article describes an action rule induction algorithm based on a sequential covering approach. Two variants of the algorithm are presented. The algorithm allows the action rule induction from a source and a target decision class point of view. The application of rule quality measures enables the induction of action rules that meet various quality criteria. The article also presents a method for recommendation induction. The recommendations indicate the actions to be taken to move a given test example, representing the source class, to the target one. The recommendation method is based on a set of induced action rules. The experimental part of the article presents the results of the algorithm operation on sixteen data sets. As a result of the conducted research the Ac-Rules package was made available.
\end{abstract}


%\begin{keyword}
%\texttt{elsarticle.cls}\sep \LaTeX\sep Elsevier \sep template
%\MSC[2010] 00-01\sep  99-00
%\end{keyword}

\end{frontmatter}

%\linenumbers

\section{Introduction}
In recent years, data mining and knowledge discovery methods have been used widely in many areas of human activity. Depending on the analytical method used, we obtain various forms of knowledge representation that have been discovered based on data. Trees and rules are representations that are considered to be the closest to how human knowledge is recorded. For this reason, tree and rule induction algorithms are most commonly used in knowledge discovery where one of the crucial features of a data model is its comprehensibility.

Decision rules are defined for descriptive and classification purposes. From a descriptive perspective, the most interesting is a set of rules representing nontrivial and useful dependencies. The comprehensibility of a rule-based data model is a particularly emphasized feature of rule-based classifiers \cite{fuernkranz2012foundations, bayardo, grzymala2003, stanczyk2020disc}.

Sequential covering rule induction algorithms can be used for both, predictive and descriptive purposes \cite{fuernkranz2012foundations, blaszczynski2011, lavravc2004, clark1989}. Despite of the development of increasingly sophisticated versions of those algorithms \cite{ liu2018induction, asadi2016acori, sikora2019guider}, the main principle remains unchanged and involves two phases: rule growing and rule pruning. In the former, the elementary conditions are determined and added to the rule premise. In the latter, some of these conditions are removed. Compared to other machine learning methods, rule sets obtained by the sequential covering algorithm, also known as the separate-and-conquer strategy (SnC), are characterized by good predictive and descriptive capabilities. When considering the former, superior results can be obtained using other methods, e.g. deep neural network, neural-fuzzy networks, support vector machines, or ensemble of classifiers \cite{liu2017survey, boser1992, rokach2010, siminski, scholkopf2001}, especially ensemble of rules \cite{gu2018massively}. However, data models obtained this way are less understandable than rule-based ones.

The issue of using rule-based representations in action mining is mainly related to the induction of action rules. An action rule is a special type of rule which represents a dependency showing a possible way to move examples from the so-called source decision class to another one called a target decision class \cite{ras2000, ras2005}. The source and target classes are also called undesired and desired decision classes, respectively. Action rules can be applied in many areas such as marketing \cite{ greco2007customer, ras2000}, healthcare and medicine \cite{touati2014mining}, sentiment analysis \cite{ranganathan2017action}, vindication, and industry.
So far, most action rule induction algorithms have been based on two approaches: \vspace{-2mm}
\begin{itemize}
  \item indirect action rule induction based on a pre-existing set of classification rules,
  \item direct induction of action rules that meet the minimum support and confidence criteria (the approach uses frequent set mining).
\end{itemize}
\vspace{-2mm}

Generally, in the field of action mining \cite{geffner1998modeling, wang2006, he2005}, two categories can be distinguished \cite{he2005}. The first category involves action induction (in particular, action rules); this category concerns the phase of knowledge discovery. The second one includes so-called transductive methods, which find the optimal recommendation for a given an example. In this paper, we work on both categories. The main contribution of our study involves:\vspace{-2mm}
\begin{itemize}
  \item proposing a sequential covering algorithm for direct induction of action rules; the algorithm generates rules starting from the source or target class, in the algorithm, different rule quality measures can be used to control the rule induction process,
  \item proposing an algorithm for recommendation discovery (i.e., determining,  for a given an example, what changes of attribute values are required to move the example from the source class to the target one); the recommendation algorithm is based on a set of induced action rules.
\end{itemize}
\vspace{-2mm}
The efficiency of the proposed algorithms is examined based on several data sets. An additional effect of the conducted research involves a software package containing implementations of the proposed methods. To our best knowledge, this is the first publicly available software package allowing for the induction of action rules and recommendation generation on this basis.

The article is organized as follows. The rest of this section contains a short literature review concerning action rule induction. In the Methods section all proposed algorithms are described, while the Results section contains a description of the experimental studies performed. At the end of the article, conclusions and a proposal for future work are presented.

\section{Related work}

The first approach in the field of action rule induction focused on generating action rules based on the existing classification, or association rule set \cite{ras2000, ras2008mining, tsay2005action, ras2005, dardzinska2012, ras2007aras}. For example, the DEAR system \cite{ras2008mining, ras2008mining, tsay2005action} generates action rules from pairs of association rules. The next proposed methods included algorithms for direct induction of action rules based on apriori-like \cite{agrawal1994} approaches \cite{he2005mining, im2008, ras2008association} or other heuristic strategies  \cite{rauch2009action, matyszok2018bidirectional, ras2008action, yang2002mining}. Most of the aforementioned approaches are based on the assumption that all possible rules are generated which meet the minimum support and confidence constraints. In the paper by Im  et al. \cite{im2010action} an agglomerative strategy for discovering action rules from an incomplete (i.e., containing missing values) information systems was presented.

Action rules reflect recommendations on the changes of attribute values; they do not indicate what operations cause the changes (e.g., the recommendation “change blood sugar level from 95 to 80” does not show what kind of medicine should be taken to fulfil this recommendation). In this case, the usability of action rules is understood as the analysis and identification of operations that should be undertaken to change the values of the attributes occurring in the premises of action rules. Such operations are called meta-actions. The analysis of dependencies between action rules and meta-actions was presented in \cite{touati2014mining, tzacheva2010association}.

Meta-action is understood as a specific action, set of actions or sequence of actions to be taken to change values of a given example to move it from the source to the target decision class. This issue is discussed in \cite{almardini2015, trepos2013, yang2002mining, hajja2014}. For example, Almardini et al. \cite{almardini2015} described procedure paths as a sequence of procedures that a given patient undertakes to reach the desired treatment. Moreover, in \cite{trepos2013} the DAKAR algorithm was proposed. The algorithm generates recommendations based on a pre-existing set of classification rules. Since the clique finding algorithm in the rule set is used for recommendation searching, the search for a good solution may be very time-consuming.

Action rule-related studies have been also carried out by Greco et al. \cite{greco2005, slowinski2005, greco2007customer}. The authors treat each classification rule as a possible intervention strategy. The meaning of the intervention strategy is explained by the following example. Let us assume the following rule: „if component A occurs in the patient’s blood, then the patient is healthy.” The rule suggests the injection of component A into patients who do not have that component in their blood. The works \cite{greco2005, slowinski2005} concentrate on a methodology for quantifying the impact of a strategy of intervention based on a decision rule quality.

In most of the above-quoted papers, the authors illustrate the efficiency of their algorithms in few data sets only. Moreover, the described experiments contain mainly the analysis of a few rules from among the generated ones. Besides, in the mentioned papers, there is no comprehensive information regarding the quality of the induced rules in terms of criteria, such as the number of induced rules, the average number of conditions in rule premises, average number of actions, average rule quality, etc.

In recent years, research on action rule induction from massive data \cite{tzacheva2017discovery, bagavathi2017sargs, ranganathan2017action, tzacheva2016mr} has also been carried out. These works focus on modifying the existing action rule induction algorithms to execute them in a distributed environment such as Hadoop MapReduce or Spark.

At the end of this review the application of decision tree induction methods to the action mining problem should be mentioned \cite{yang2007extracting, alam2012actionable, subramani2016mining, yang2003postprocessing, ling2002mining}. Works \cite{yang2007extracting, ling2002mining} depicted transductive action mining methods. These papers describe decision tree induction methods and optimal recommendation discovery based on the induced trees. Articles \cite{alam2012actionable, subramani2016mining, yang2003postprocessing} focus on the post-processing of decision trees to discover the optimal recommendation. Decision tree ensembles have also been applied to the action mining problem \cite{cui2015optimal, tolomei2017interpretable}.

\section{Methods}

\subsection{Basic notions}
Let $E(A, \{d\})$ be a data set of $|E|$ examples (observations, instances), each being characterized by a set of conditional attributes $A=\{a_1, a_2,...,a_{|A|}\}$ and a decision attribute $d$. Conditional attributes can be of symbolic (discrete-valued), ordinal or numeric (real-valued) type. Each training example $x \in E$ can be represented as a vector $x=(x_1, x_2, \ldots, x_{|A|}, d(x))$ where $a_i(x)=x_i$ for each $i\in \{1,2,\ldots,|A|\}$. The decision attribute is of symbolic type, it corresponds to a discrete class identifier, i.e., for $x \in E$, $d(x) \in \{C_1, C_2, ...,C_l\}$.

In action rule induction, it is assumed that conditional attributes are differentiated as stable and flexible. An attribute is defined as stable if the values of the attribute assigned to examples cannot be changed. Otherwise, the attribute is defined as flexible.

Let $R$ be a set of classification rules generated by the induction algorithm, referred later as a rule-based data model or, simply, a model. Each rule $r \in R$ has the form:
\\
\\
$
\IF w_1 \AND w_2 \AND \ldots \AND w_n \THEN C
$
\\
\\
The premise of a rule is a conjunction of elementary conditions $w_i\equiv a_i \odot x_i$, with $x_i$ being an element of the $a_j$ domain and $\odot$ representing a relation ($=$ for symbolic attributes; $<, \leq, >, \geq$ for ordinal and numerical ones). The value $C$ in the rule conclusion indicates one of the decision class identifiers $\{C_1, C_2, ...,C_l\}$. The meaning of a rule is as follows: if an example fulfils all conditions specified in the conditional part, then it belongs to the decision class specified in the rule conclusion. Additionally, an example satisfying the conditions specified in the rule premise is stated to be covered by the rule.

\subsection{Separate-and-conquer classification rule induction}

The presented algorithm induces rules according to the separate-and-conquer principle \cite{furnkranz1999, michalski1973discovering}. An important factor determining the performance and comprehensibility of the resulting rule-based model is a selection of a rule quality measure \cite{bruha1997, an2001rule, wrobel2016} (rule learning heuristic \cite{furnkranz2005, janssen2010quest, minnaert}) that supervises the rule induction process. Let $r$ be the considered classification rule. The examples whose labels are the same as the conclusion of $r$ will be referred to as positive, while the others will be called negative. The confusion matrix for a rule consists of the number of positive and negative examples in the entire training set ($P$ and $N$), and the number of positive and negative examples covered by the rule ($p$ and $n$). The idea can be straightforwardly generalized for weighted examples by replacing numbers of examples in the confusion matrix by sums of their weights.
Based on the rule confusion matrix many rule quality measures are defined. In our research the following rule quality measures were considered: C2 \cite{bruha1997}, Correlation \cite{furnkranz2005}, Lift \cite{bayardo}, RSS (Rule Specificity and Sensitivity) \cite{sikora2013data}, wLap (weighted Laplace). These measures evaluate rules using various criteria resulting in very different models. For instance, RSS (also known as WRA (Weighted Relative Accuracy) \cite{furnkranz2005}) considers equally sensitivity ($p/P$) and specificity ($1 - n/N$) of the rule according to the formula $\textrm{RSS} = p/P - n/N$. The wLap defined as $(p+1)(P+N)/((p+n+2)P)$ revises the rule precision calculates based on the entire training set. The C2 measure (\ref{c2}) is a multiplication of modified rule precision ($p/(p+n)$) and modified rule coverage ($p/P$). In addition, the evaluation of the rule according to wLap and C2 measures considers the distribution of the number of positive and negative examples.

\begin{equation}\label{c2}
{\left(\frac{Np-Pn}{N(p+n)}\right)\left(\frac{P+p}{2P}\right).}
\end{equation}

Another common measure is the Gain measure which measures entropy of an outcome variable $Y$ given random variable $X$ as:
\begin{equation}
H(Y|X) = -\sum_{x \in X} P(x) \sum_{y \in Y} P(y|x) \log{P(y|x).}
\end{equation}
In our case $Y$ indicates class (positive/negative) and $X$ denotes whether the rule covers the example (covered/uncovered). Therefore,
\begin{align}
 & P(X=\textrm{covered}) = \ (p + n)/(P + N), \\
 & P(Y=\textrm{positive}\ |\ X=\textrm{covered}) =\ p / (p + n), \\
 & P(Y=\textrm{positive}\ |\ X=\textrm{uncovered}) =\ (P - p) / (P + N - p - n).
 \end{align}
The opposite probabilities, i.e., $P(X=\textrm{uncovered})$, $P(Y=\textrm{negative}\ |\ X=\textrm{covered})$, and $P(Y=\textrm{negative}\ |\ X=\textrm{uncovered})$ can be calculated straightforwardly by subtracting from 1 appropriate value.

\begin{algorithm}[!b]
\begin{spacing}{1.5}
	\small
	\begin{algorithmic}[1]
		\caption{Separate-and-conquer rule induction.}
		\label{alg:conquer}
		
		\Require
		$E(A,\{d\})$---training data set,
		\mincov---minimum number of yet uncovered examples that a new rule has to cover.
		\Ensure $R$---rule set.
		\State $E_{U} \gets E$	\Comment{set of uncovered examples}
		\State $R \gets \emptyset$ \Comment{start from an empty rule set}
		\Repeat
		\State $r \gets \emptyset$ \Comment{start from an empty premise}
		\State $r \gets \Call{Grow}{r, E, E_{U}, \mincov}$ \Comment{grow conditions}
		\State $r \gets \Call{Prune}{r,E}$ \Comment{prune conditions}
		\State $R \gets R \cup \{r\}$
		\State $E_{U} \gets E_{U}\setminus\Call{Cov}{r, E_U}$ \Comment{remove from $E_U$ examples covered by $r$}
		\Until{$|E_{U}| < \mincov$}	
	\end{algorithmic}
\end{spacing}
\end{algorithm}

\begin{algorithm}[!t]
\begin{spacing}{1.5}
	\begin{algorithmic}[1]
		\caption{Growing a rule.}
		\label{alg:grow}
		\Require
		$r$---input rule,
		$E$---training data set,
		$E_{U}$---set of uncovered examples,
		\mincov---minimum number of previously uncovered examples that a new rule has to cover.
		\Ensure
		$r$---grown rule.
		
		\Function{Grow}{$r$, $E$, $E_U$, $mincov$}
		
		\Repeat \Comment{iteratively add conditions}
		\State $w_\textrm{best} \gets \emptyset$ \Comment{current best condition}
		\State $q_\textrm{best} \gets -\infty,\quad \textrm{cov}_\textrm{best} \gets -\infty$ \Comment{best quality and coverage}
		
		\State $E_{r} \gets$ \Call{Cov}{$r$, $E$} \Comment{examples from $E$ satisfying $r$ premise}
		
		\For{$w \in$ \Call{GetPossibleConditions}{$E_r$}}
			\State $r_w \gets r \AND w$ \Comment{rule extended with condition $w$}
			\State $E_{r_w} \gets \Call{Cov}{r_w, E}$
			\If {$|E_{r_w} \cap E_U| \geq \mincov$} \Comment{verify coverage requirement}
				\State $q \gets$ \Call{Quality}{$E_{r_w}$, $E \setminus E_{r_w}$} \Comment{rule quality measure}
		
				\If {$q > q_\textrm{best}$ \textbf{or} ($q = q_\textrm{best}$ \textbf{and} $|E_{r_w}| > \textrm{cov}_\textrm{best}$)}

					\State $w_\textrm{best} \gets w,\quad q_\textrm{best} \gets q,\quad \textrm{cov}_\textrm{best} \gets |E_{r_w}|$
				
				\EndIf
			\EndIf
		\EndFor		
        \Statex
		\State $r \gets r \AND w_\textrm{best}$
		\Until{$w_\textrm{best} = \emptyset$}

		\State \Return{$r$}
		\EndFunction
	\end{algorithmic}
\end{spacing}
\end{algorithm}

Separate-and-conquer top-down rule induction heuristic \cite{fuernkranz2012foundations} adds rules iteratively to the initially empty set as long as the entire data set becomes covered (Algorithm~\ref{alg:conquer}). Vvery rule must cover at least $\mincov$ previously uncovered examples to ensure the convergence. Therefore, the generation of consecutive rules stops when there are less than $\mincov$ uncovered examples left. The induction of a single rule consists of two stages: growing and pruning. In the former (presented in Algorithm~\ref{alg:grow}), elementary conditions are added to the initially empty rule premise. When extending the premise, the algorithm considers all possible conditions built upon all attributes (line 6: \textsc{GetPossibleConditions} function call), and selects those leading to the rule of the highest quality (lines 10--12). In the case of nominal attributes, conditions in the form $a_i = x_i$ for all values $x_i$ from the attribute domain are considered. For continuous attributes, $x_i$ values that appear in the observations covered by the rule are sorted. Then, the possible split points $x_i$ are determined as arithmetic means of subsequent $a_i$ values and conditions $a_i < x_i$ and $a_i \geq x_i$ are evaluated. If several conditions render the same results, the one covering more examples is chosen. Pruning can be considered the opposite of growing. It iteratively removes conditions from the premise, each time making an elimination leading to the largest improvement in the rule quality. The procedure stops when no conditions can be deleted without decreasing the quality of the rule or when the rule contains only one condition. Finally, for comprehensibility, the rule is post-processed by merging conditions based on the same numerical attributes. E.g., conjunction $a_i \geq 3 \AND a_i \geq 5 \AND a_i < 10$ will be presented as $a_i \in [5,10)$.

Rule sets induced by the separate-and-conquer heuristic are unordered. Therefore, applying the induced rule set $R$ (rule-based data model) to the classification problem requires evaluating set $R_\textrm{cov} \subseteq R$ of rules covering a classified example and aggregating the results. This differs from ordered rule sets (decision lists), where the first rule covering the investigated example determines the model response. In classification, the output class label is obtained as a result of voting---each rule from $R_\textrm{cov}$ votes with its value of the quality measure this may be a different measure than the one used during the induction~\cite{wrobel2016}.

The detailed information about our version of rule induction using a separate-and-conquer approach can be found in~\cite{wrobel2016, sikora2019guider}. 	

\subsection{Action rules}
Let us consider the following formula:

\noindent $
\IF w_{1S} \rightarrow w_{1T} \AND w_{2S} \rightarrow w_{2T} \AND \ldots \AND w_{nS} \rightarrow w_{nT} \THEN C_S \rightarrow C_T
$
\\
\\
\noindent We will refer to such a formula as an action rule. Let us suppose that $r$ is an action rule. The rule $r$ may be seen as a composition of two classification rules:
\\
\\
$
r_S \equiv \IF w_{1S} \AND w_{2S} \AND \ldots \AND w_{nS} \THEN C_S,
\\
r_T \equiv \IF w_{1T} \AND w_{2T} \AND \ldots \AND w_{nT} \THEN C_T.
$
\\
\\
The first rule is called a source part of the action rule $r$. The second one is called a target part of $r$.
Additionally, the following names will be used in the subsequent part of the article: \vspace{-2mm}
\begin{itemize}%[leftmargin=*,labelsep=5.8mm]
\item   the premise of the $r_S$ ($r_T$) rule will be called a premise of the source (target) part of the action rule $r$,
\item	the decision class $C_S$ ($C_T$) in the conclusion of $r_S$ ($r_T$) will be called a source (target) decision class,
\item	the composite elementary condition $w_{iS} \rightarrow w_{iT}$ in the action rule premise will be called an elementary action,
\item	the elementary condition $w_{iS}$ ($w_{iT}$) in the elementary action will be called a source (target) of the elementary action,
\end{itemize}
\vspace{-2mm}
%tutaj zdefiniujemy zbiory: reguł akcji, reguł r_S i r_T; akcji elementarnych w regule, akcji alementranych we wszystkich regułach etc.
Let $R$ be a set of action rules. We will denote by $R_S$ ($R_T$) the set of all rules $r_S$ ($r_T$) obtained based on the rule set $R$ (i.e.
$r_S \in R_S \Leftrightarrow \exists r \in R$ such that $r_S$ is the source part of the action rule $r$, $r_T \in R_T \Leftrightarrow \exists r \in R$ such that $r_T$ is the target part of the action rule $r$).

Let $r \in R$ be an action rule. We will denote by: \vspace{-2mm}
\begin{itemize}%[leftmargin=*,labelsep=5.8mm]
\item   $W_r$ the set of all elementary actions which appear in $r$,
\item	$W_{r_S}$ the set of all elementary conditions which appear in $r_S$,
\item	$W_{r_T}$ the set of all elementary conditions which appear in $r_T$,
\end{itemize}
\vspace{-2mm}
For the set of action rules $R$ we also define sets $W_R$, $W_{R_S}$ and $W_{R_T}$ according to the following formulae: \vspace{-2mm}
\begin{itemize}%[leftmargin=*,labelsep=5.8mm]
\item   $W_R=\bigcup_{r \in R} W_r$,
\item	$W_{R_S}=\bigcup_{r \in R} W_{r_S}$,
\item	$W_{R_T}=\bigcup_{r \in R} W_{r_T}$.
\end{itemize}
\vspace{-2mm}

Let be given an elementary action $(a \odot_1 x_i \rightarrow  a \odot_2 x_j)$. We consider the action as a need (recommendation, requirement) for changing the value of attribute $a$ from present range
$(a \odot_1 x_i)$ to the range $(a \odot_2 x_j)$. For example, the elementary action  ($ body \; temperature > 38 \degree C$) $\rightarrow$ ($ body \; temperature < 37 \degree C)$) indicates the need for reducing body temperature. An action rule is interpreted as a conjunction of needs whose fulfilment will cause that the example classified to the source class $C_S$ will change its assignment and become an example representing the target class $C_T$. For example, an action rule:
\\
\\
$
\IF (( body \; temperature > 38 \degree C) \rightarrow ( body \; temperature < 37 \degree C)) \AND (( body \; temperature > 38^{0}C) \rightarrow ( body \; temperature > 35 \degree C)) \AND (( pus \; on \; tonsils = Yes) \rightarrow ( pus \; on \; tonsils = No)) \THEN ( ill \; = Yes) \rightarrow ( ill \; = No)
$
\\
\\
informs that normalizing body temperature and elimination of pus from tonsils will cure the patient. The presented rule can be written in the comprehensive form as:
\\
\\
$
\IF (( body \; temperature > 38 \degree C) \rightarrow ( body \; temperature \in (35 \degree C, 37 \degree C))) \AND (( pus \; on \; tonsils = Yes) \rightarrow ( pus \; on \; tonsils = No)) \THEN ( ill \; = Yes) \rightarrow ( ill \; = No)
$
\\
\\
Action rules can contain two special types of elementary actions. In the first, only the source of elementary action is specified. For example, a rule
\\
\\
$\IF$ ($ body \; temperature > 38 \degree C$) $\AND$ (($ pus \; on \; tonsils = Yes$) $\rightarrow$ ($ pus \; on \; tonsils = No$)) $\THEN$ ($ ill \; = Yes$) $\rightarrow$ ($ ill \; = No$)
\\
\\
shows that if we remove pus from tonsils for  patients with a $body \; temperature > 38^{0}C$, they will recover. The condition ($ body \; temperature > 38^{0}C$) is here the standard elementary condition constraining the set of patients for whom the elemental action ($ pus \; on \; tonsils = Yes$) $\rightarrow$
($ pus \; on \; tonsils = No$) is applicable. Such conditions in the premises of action rules we will simply call constraints. Each constraint in the premise of the action rule r is an element of the set $W_{r_S}$. An example of the second special type of an elementary action is the following one:
\\
\\
 $(\rightarrow$ ($ body \; temperature \in (35^{0}C, 37^{0}C)$) (which is a comprehensive form of
 $(\rightarrow$ ($ body \; temperature < 37^{0}C$)) $\AND$ $(\rightarrow$ ($body \; temperature > 35^{0}C)$).
\\
\\
We will call this action a narrowing one. The narrowing action indicates a need for narrowing the attribute value, regardless of its current value, to the set indicated in its target part. The condition in the narrowing action is an element of the set $W_{r_T}$.

By analogy to $p, n, P, N$ values used for quality evaluation of classification rules, we will use the following notations for action rules:
\begin{itemize}
\item $p_S$ - the number of positive examples covered by the $r_{S}$ rule,
\item $n_S$ - the number of negative examples covered by the $r_{S}$ rule,
\item $p_T$ - the number of positive examples covered by the $r_{T}$ rule,
\item $n_T$ - the number of negative examples covered by the $r_{T}$ rule.
\end{itemize}

\subsection{Separate-and-conquer action rule induction}

\begin{algorithm}[b]
\begin{spacing}{1.5}
%	\small
	\begin{algorithmic}[1]
		\caption{Action rule growing.}
		\label{alg:grow_action_rule}
		
		\Require
		$r$---input action rule,
		$E$---training data set,
		$E_{U}$---set of examples uncovered by source part of $r$,
		\mincov---minimum number of previously uncovered examples that a new rule has to cover.
		\Ensure
		$r$---grown rule.

		\Function{GrowActionRule}{$r$, $E$, $E_U$, $mincov$}

		\State $r_S \gets \Call{GetSourcePart}{r}$
		\State $r_T \gets \Call{GetTargetPart}{r}$
		
		\State $q_{\textrm{best}_S} \gets -\infty,\quad \textrm{cov}_{\textrm{best}_S} \gets -\infty$ \Comment{best quality and coverage of source part}
		\State $q_T \gets -\infty,\quad \textrm{cov}_T \gets -\infty$ \Comment{best quality and coverage of target part}
		
		\Repeat
		\State $w_{\textrm{best}_S} \gets \emptyset$ \Comment{current source best condition}
		\State $w_T \gets \emptyset$ \Comment{current target condition}
		\State $E_{r} \gets$ \Call{Cov}{$r_S$, $E$} \Comment{examples from $E$ satisfying $r_S$ premise}
		
		\For{$w \in$ \Call{GetPossibleConditions}{$E_r$}}
			\State $r_{S_w} \gets r_S \AND w$ \Comment{source rule extended with condition $w$}
			\State $E_{r_{S_w}} \gets \Call{Cov}{r_{S_w}, E}$
			\If {$|E_{r_{S_w}} \cap E_U| \geq \mincov$} \Comment{verify coverage requirement}
				\State $q \gets$ \Call{Quality}{$E_{r_{S_w}}$, $E \setminus E_{r_{S_w}}$} \Comment{rule quality measure}
		
				\If {$q > q_{\textrm{best}_S}$ \textbf{or} ($q = q_{\textrm{best}_S}$ \textbf{and} $|E_{r_S{_w}}| > \textrm{cov}_{\textrm{best}_S}$)}
					\State $w_{\textrm{best}_S} \gets w,\quad q_{\textrm{best}_S} \gets q,\quad \textrm{cov}_{\textrm{best}_S} \gets |E_{r_{S_w}}|$
				\EndIf
		
			\EndIf
		\EndFor
        \Statex

		\State $E_{r} \gets$ \Call{Cov}{$r_T$, $E$} \Comment{examples from $E$ satisfying $r_T$ premise}
		\State $a \gets$ \Call{GetAttribute}{$w_{\textrm{best}_S}$}
		\For{$w \in$ \Call{GetPossibleConditionsForAttribute}{$E_r$,$a$}}
			
			\State $r_{T_w} \gets r_T \AND w$
			\State $E_{r_T{_w}} \gets \Call{Cov}{r_{T_w}, E}$
			
			\If {$|E_{r_T{_w}}| \geq \mincov$} \Comment{verify coverage requirement}
			
				\State $q \gets$ \Call{Quality}{$E_{r_T{_w}}$, $E \setminus E_{r_T{_w}}$} \Comment{rule quality measure}
		
				\If {$q > q_T$ \textbf{or} ($q = q_T$ \textbf{and} $|E_{r_T{_w}}| > \textrm{cov}_T$)}
					\State $w_T \gets w,\quad q_T \gets q,\quad \textrm{cov}_T \gets |E_{r_T{_w}}|$
				\EndIf
		
			\EndIf
			
		\EndFor
        \Statex

		\State $r \gets r \AND (w_{\textrm{best}_S} \rightarrow w_T)$
		\Comment{Extend rule with new elementary action}

		\Until{$w_{\textrm{best}_S} = \emptyset$}
		
		\State \Return{$r$}
		
		\EndFunction	
	\end{algorithmic}
\end{spacing}
\end{algorithm}

The induction of action rules is similar to the induction of classification ones. Action rule induction starts from the definition of the conclusion of an action rule. Suppose an action rule indicating the transition from class $C_S$ to $C_T$ is being induced. The conclusion of such a rule has the form  $C_S \rightarrow C_T$.

In the premise of an action rule there are elementary actions. Elementary actions are added, one by one, to the premise according to the rule growing strategy described in Algorithm 2. Adding a new elementary action $w_{i_S} \rightarrow w_{i_T}$ requires specifying both the source ($w_{i_S}$) and the target ($w_{i_T}$) of the elementary action. The order in which the elementary conditions $w_{i_S}$ and $w_{i_T}$ are specified depends on the version of the separate-and-conquer action rule induction algorithm. There are two versions of our algorithm, the Forward and the Backward.

In the Forward method, while generating consecutive elementary actions  $w_{i_S} \rightarrow w_{i_T}$, the source $w_{i_S}$ of the action is induced first and next the target part $w_{i_T}$ of the action is determined. In the Backward method, the reverse approach is applied, first the $w_{i_T}$ condition is induced, and then $w_{i_S}$.

The main difference between the two approaches lies in how they search for the best attribute at a given stage of the rule growing phase - in discrimination of examples from $C_S$ and $C_T$ classes.

In the Forward method, the algorithm searches for the best attribute for the source part $w_{i_S}$ of an elementary action. It means that the algorithm tries to find the best elementary condition for the classification rule $r_S$. Let us assume that this attribute is $a$, and the source part $w_{i_S}$ of the elementary action is already induced (based on attribute $a$). The target part $w_{i_T}$ of the elementary action is also built based on the already chosen attribute $a$.

In the Backward method, it is precisely the opposite. The algorithm looks for the best attribute for the target part $w_{i_T}$ the elementary action (i.e. best elementary condition for the classification rule $r_T$), and then the source part $w_{i_S}$ of the elementary action (for the already found attribute) is induced.

In the proposed approach different measures of rule quality may supervise the process of induction of the source ($w_{i_S}, i \in  \{1, ..., n\}$) and target ($w_{i_T}, i \in \{1, ..., n\}$) parts of elementary actions.

The Forward approach carries out the induction of action rules from the source class while the Backward one from the target class point of view, respectively. Mixed strategies are also possible, but this article does not consider such an approach.

There are also some differences between the induction of classification rules and action rules at the pruning phase.
As it is known, pruning involves the removal of elementary conditions. In the case of action rules, this is the removal of elementary actions. Action rule pruning is performed in both strategies (Forward and Backward) identically.
An elemental action $w_{i_S} \rightarrow w_{i_T}$ consists of two conditions  $w_{i_S}$, $w_{i_T}$. Removing an elementary action consists in:
\begin{itemize}
\item removing the source part $w_{i_S}$ of the elementary action $w_{i_S} \rightarrow w_{i_T}$ and checking whether the rule $r_S$  with the removed condition $w_{i_S}$ has quality not worse than the rule containing $w_{i_S}$; if this is the case, the condition is removed and the elementary action takes the form $\rightarrow w_{i_T}$,
\item removing the target part $w_{i_T}$ of the elementary action $w_{i_S} \rightarrow w_{i_T}$  and checking whether the rule $r_T$  with the removed condition $w_{i_T}$ has quality not worse than the rule containing the removed one; if so, then the $w_{i_T}$ condition is removed from the elementary action, and the elementary action takes the form $w_{i_S}$ (that is, the elementary action becomes the constraint and limits the set of examples covered by the pruned action rule,
\item if in both cases the removal of the $w_{i_S}$ condition causes no deterioration in the quality of the $r_S$ rule and the removal of the $w_{i_T}$ condition causes no deterioration in the quality of the $r_T$ rule, then the entire elementary action $w_{i_S} \rightarrow w_{i_T}$ is removed from the rule premise.
\end{itemize}

Algorithm 3 presents a modification of Algorithm 2 illustrating the growing phase of an action rule. It is a part of the Forward version of the algorithm. During the growing phase, the action rule is perceived as two classification rules, $r_S$ and $r_T$ (Algorithm \ref{alg:grow_action_rule}, lines 2-3).

The best attribute for the source part of an elementary action $w_{best_S}$ being induced is selected with the use of rule quality measure (e.g., $C2$) by temporary extension of the premise of $r_S$ with candidate condition ($w$) and evaluation the quality of such an extended rule (Algorithm \ref{alg:grow_action_rule}, lines 10-16). The target part of the elementary action is searched based on the set of all possible elementary conditions for the attribute already chosen during the source part of the elementary action induction (Algorithm \ref{alg:grow_action_rule}, lines 18-25). The procedure is repeated until no more source parts of elementary actions increasing the quality of the $r_S$ rule can be found or the minimal coverage criterion is no longer met.
\\

\noindent \textbf{Example}

\noindent To illustrate how the algorithm works, we present an illustrative example of the induction of one action rule based on a well-known Monk1 data set. All attributes in this set are symbolic and the target class (denoted as 1) is defined as follows: all examples meeting the condition $a_1 = a_2$ or $a_5 = 1$ are labelled as belonging to the target class. The considered version of the rule induction algorithm does not allow the induction of conditions $a_1 = a_2$, but let us analyse the growing phase of an exemplary action rule according to the Forward method.

The rule has initially empty premise and conclusion $(class=0 \rightarrow class=1)$. The process of elementary actions adding to the rule premise is illustrated in table \ref{table:1}. The table contains the best conditions found in consecutive iterations of the rule growing phase and qualities of the rules $r_S$ and $r_T$ extended with $w_{\textrm{best}_S}$ and $w_T$, respectively. As a quality measure rule precision was used.

\begin{table}
\centering
\begin{tabular}{ c c c c c }
\hline
 iteration & $w_{\textrm{best}_S}$ & $q_{\textrm{r}_S}$ &  $w_T$ & $q_{\textrm{r}_T}$ \\
\hline \hline
 1 & $(a_1 = 1)$ &  0.69 & $(a_1 = 3)$  & 0.70 \\
 2 & $(a_2 = 2)$ & 0.88  & $(a_2 = 3)$  &  1.00 \\
 3 & $(a_6 = 2)$ & 0.90 & $(a_6 = 2)$ & 1.00 \\
\hline
\end{tabular}
\caption{Consecutive source and target parts of elementary actions induced during the action rule growing (q - rule precision)}
\label{table:1}
\end{table}

\noindent After the rule gowning phase, the action rule has the following form:
\\

\noindent $\IF ((a_1=1) \rightarrow (a_1=3)) \AND ((a_2=2) \rightarrow (a_2=3)) \AND ((a_6=2) \rightarrow (a_6=2)) \THEN (class=0) \rightarrow (class=1)$.
\\

The rule is characterised by the following statistics: $p_S = 9, n_S = 1, p_T = 17, n_T = 0$

After the growing phase, the rule pruning phase is invoked. The selected stages of removing elementary actions from the rule premise are illustrated in table \ref{table:2}. The table contains successive forms of the pruned rule and qualities of their rules $r_S$ and $r_T$. The RSS measure was used during the rule pruning process.

\begin{table}
\centering
\scalebox{0.9} {
\begin{tabular}{ c c c }
\hline
 rule premise & $q_{\textrm{r}_S}$ & $q_{\textrm{r}_T}$  \\
\hline \hline
 $(a_1=1) \rightarrow (a_1=3) \AND (a_2=2) \rightarrow (a_2=3)) \AND (a_6=2) \rightarrow (a_6=2)$ & 0.13 & 0.27   \\
 $(a_1=1) \rightarrow (a_1=3)) \AND ((a_2=2) \rightarrow (a_2=3)) \AND (a_6=2) \rightarrow $ & 0.13 & 0.27   \\
 $(a_1=1) \rightarrow (a_1=3) \AND (a_2=2) \rightarrow (a_2=3) $ & 0.21 & 0.27   \\
 $(a_1=1) \rightarrow (a_1=3) \AND ((a_2=2) \rightarrow $ & 0.21 & 0.24   \\
 $(a_1=1) \rightarrow (a_1=3) \AND \rightarrow (a_2=3) $ & 0.26 & 0.27   \\
\hline
\end{tabular}
}
\caption{Illustration of the action rule pruning (q - RSS)}
\label{table:2}
\end{table}

Finally, after rule pruning the pruned rule has the following form:
\\
\\
$\IF ((a_1=1) \rightarrow (a_1=3)) \AND (\rightarrow (a_2=3)) \THEN (class, 0) \rightarrow (class, 1)$
\\

The pruned rule is characterised by the following statistics: $p_S = 31, n_S = 14, p_T = 17, n_T = 0$

The rule recommends setting values of both attributes $a_1$ and $a_2$ to $3$. Such changes assure that all examples covered by $r_S$ will represent the target decision class. This recommendation is consistent with the target class definition for the Monk1 data set. $\blacksquare$
%\end{Example}
\\

The action rule induction algorithm can be run for each pair of decision classes or a given decision class (the target one) and all remaining decision classes joined into one decision class (the source decision class).
%tutaj dopisć kwestię postprocessingu reguł czyli jesli wystepuje klka akcji elementarnych zbudowanych na podstawie tego samego atrybutu to robimy AND

\subsection{Recommendation induction -- resolving conflicts within a set of action rules}

The set of induced rules represents a new, discovered knowledge. Each rule $r_S \rightarrow r_T$ shows what changes of the attribute values appearing in the premise of $r_S$ are necessary to change the classification of an example from class $S$ to $T$. Many action rules can cover new, unseen examples representing the source class $S$. In such a situation, the crucial question is: ,,what elementary actions should be fulfilled to move the example to the area of the feature space covered by examples representing the target class'' This problem can be solved by inducing a special type of rule (or rules - if we are interested in more than one recommendation) in a new data set. The new data set is generated based on the induced action rules. We will call this set a set of meta-examples.

The set of meta-examples is defined as follows. Suppose a set $R$ of action rules is available. Let us assume that a set of examples $E = (A, \{d\})$ is given, the meta-table $mE=(mA, \{\})$ is composed of the set of meta-attributes $mA = \{\prescript{}{m}a_{i_1}, \prescript{}{m}a_{i_2}, ..., \prescript{}{m}a_{i_m}\}$.
If an attribute $a \in A$  is not an element of any elementary action among the rules from $R$, then $a$ does not belong to the set of meta-attributes $mA$. In other words, if $a \in A$ and in the set $W_R$ there is not an elementary action build based on $a$ then $\prescript{}{m}a \notin mA$

If the attribute $a \in A$ occurs in at least one elementary action among the rules from $R$ and the attribute is of symbolic type, then $\prescript{}{m}a:=a$.

%Each rule $r \in R$ has a form $r_S \rightarrow r_T$. For our considerations it is not important which decision class is the source and which is the target one.
Let $a \in A$ be a numeric type attribute and $a$ occurs in at least one elementary action among the rules from $R$. %The following transformation procedure from $a$ to $\prescript{}{m}a$ is applied:
The value set of the meta-attribute $\prescript{}{m}a$ is determined on basis of the following assumptions and transformations:
\begin{itemize}
\item by $W_{r_{a_ S}}$ ($W_{r_{a_ T}}$) we denote the set of all elementary conditions of rules $r_S$ ($r_T$) containing the attribute $a$;
\item let $W_a = W_{r_{a_ S}} \cup W_{r_{a_ T}}$; the set $W_a$ includes all elementary conditions built on the basis of the attribute $a$;
\item each elementary condition takes the form $a > v$ or $a <= v$, where $v$ is a certain value from the domain of $a$;
\item sorting all values $v$ appearing in $W_a$, makes it possible to define a partition (discretisation) of the value set of $a$;
\item values of $\prescript{}{m}a$ are defined as consecutive identifiers (natural numbers) of the elements of discretisation of $a$.
\end{itemize}

For each real type attribute $a \in A$, the meta-attribute $\prescript{}{m}a$ is an ordinal type attribute in $mA$.

\noindent \textbf{Example}
\
\noindent Let $W_a = \{ a \leq 3, a > 5, a > 7, a \leq 6 \}$. On this basis, we obtain the following partition of the value set of $a$:\\
\centerline{$(min_a, 4], (3, 5], (5, 6], (6, 7], (7, max_a)$}
\noindent The values $min_a$ and $max_a$ are the minimal and maximal values of $a$ in $E$. This partition defines the value set of $\prescript{}{m}a$, the set has 5 values, e.g. the interval $(3.5]$ is assigned to the value $2$ of $\prescript{}{m}a$. $\blacksquare$ \\

%\end{Example}

Let us denote by $\prescript{}{m}V_{a_i}$ the set of values of the meta-attribute $\prescript{}{m}a_i$. Elements of the meta-table (meta-examples) are m-tuples belonging to the Cartesian product
$\prescript{}{m}V_{a_1} \times  \prescript{}{m}V_{a_2} \times  ... \times  \prescript{}{m}V_{a_m}$.
The decision attribute in the meta-table does not exist (more precisely, it is not defined because it is unimportant for further considerations).

\noindent \textbf{Example}
\\
\noindent Suppose there are two attributes in $A$. A numeric attribute and a symbolic one. The numeric attribute values are divided as in the previous example, and the symbolic attribute has only two values $Yes$ and $No$.
There are $5*2 $ meta-examples in the meta-table. For example in the meta-table there are the following two examples:\\
\centerline{$1, Yes$}
\centerline{$4, No$}
\noindent where $1$ indicates the interval $(min_1, 3]$, and $4$ the interval $(6, 7]$. $\blacksquare$ \\
%\end{Example}

\noindent \textbf{Example}
\\
\noindent Let us assume a simple set $R$ containing only two below action rules is given:\\
$
\textrm{r1:} \IF (( body \; temperature>38 \degree C) \rightarrow ( body \; temperature<36.6 \degree C)) \AND (( pus \; on \; tonsils = Yes) \rightarrow ( pus \; on \; tonsils = No)) \THEN ( ill \; = Yes) \rightarrow ( ill \; = No)\\
\textrm{r2:} \IF (( body \; temperature>37.5 \degree C) \rightarrow ( body \; temperature<37 \degree C)) \AND (( pus \; on \; tonsils = No)) \THEN ( ill \; = Yes) \rightarrow ( ill \; = No)\\
$
Based on these rules we have:
\begin{itemize}
\item $W_{body \; temperature} = \{ ( body \; temperature>38 \degree C),  ( body \; temperature>37.5 \degree C), ( body \; temperature<36.6 \degree C),  ( body \; temperature<37 \degree C)\}$
\item $W_{pus \; on \; tonsils}= \{ ( pus \; on \; tonsils = No),  ( pus \; on \; tonsils = Yes) \}$
\end{itemize}
\noindent and partitions $(min, 36.6], (36.6, 37], (37, 37.5], (37.5, 38], (38, max_a)$, {$No, Yes$}, where $min$ and $max$ are the minimal and maximal values of the \textit{body temperature} attribute, respectively.
%For practical reasons, $W_{R_S}$ ($W_{R_T}$) set is usually also extended with two special ranges: from the lowest value possible (infimum) of given attribute' value range to lowest value present in $W_{R_S}$ ($W_{R_T}$), and from highest value present in $W_{R_S}$ ($W_{R_T}$) to the supremum of the %value range. Often the infimum and suprememum are simply negative and positive infinity.

A meta-table defined on the basis of $R$ is presented in table \ref{table:3}. $\blacksquare$ \\

\begin{table}
\centering
\begin{tabular}{  c c }
\hline
$body \; temperature$ & $ pus \; on \; tonsils$\\
\hline \hline
$1$ & $No$ \\
$1$ & $Yes$ \\
$2$ & $No$ \\
$2$ & $Yes$ \\
$3$ & $No$ \\
$3$ & $Yes$ \\
$4$ & $No$ \\
$4$ &  $Yes$ \\
$5$ & $No$ \\
$5$ &  $Yes$ \\
\hline
\end{tabular}
\caption{Example meta-table}
\label{table:3}
\end{table}
%\end{Example}

According to the above definitions, each example in a meta-table covers a certain part/area in the attribute space, and thus is covered by several examples from the original table $E=(A, \{d\})$.

Some of the meta-examples in the meta-table are covered by the rules from the set $R_S$, some by the rules from $R_T$, and some by both types of these rules. Some meta-examples are not covered by any rules.

The principle of the recommendation induction algorithm working is as follows. For a given test example belonging to the source class $C_S$ a meta-example covering the largest possible number of examples from $C_T$ class and the smallest possible number of examples from $C_S$ class is searched. Let us also note that every test example is covered by one meta-example only.

Thus for a test example, the recommendation is a guideline concerning the changes of attribute values of the test example to make it covered by a meta-example that fulfils the above principle. To reduce the number of necessary changes of values of attributes, the recommendation induction algorithm heuristically searches for only such changes
that ensure the change of the example assignment form the $C_S$ class to the $C_T$ class. For this purpose, the rule induction algorithm in the meta-table is run. As there is not defined a decision attribute in the meta-table, the algorithm requires indicating the target class $C_T$ to which the test example should be moved (e.g., a regular customer). All remaining classes represent the source class unless the source class of the test example is known (e.g. an ordinary customer). A slightly modified version of the classification rule induction algorithm (Algorithm 2) is used to find recommendations.

For a given test example $x$ and target class $T$, all examples from $E = (A, \{d\})$ representing the target class are considered as positive examples. All remaining examples are considered negative. During recommendation induction elementary conditions are searched in $E = (A, \{d\})$ but the  evaluation of the quality of rule (i.e. recommendation) being induced is calculated on the basis of $E = (A, \{d\})$. Rule induction is carried out only for the target class $T$. Each induced rule represents one recommendation for the example $x$. As the final recommendation, the recommendation with the highest quality is chosen.\\

%tutaj trzeba wyjasnić diałanie rekomendacji
%The algorithm may induce many recommendations, but according to the sequential covering approach, usually the first induced recommendation (an action rule induced in the meta-table) is the best one.

\noindent \textbf{Example}
\\
\noindent Following the previous example.
Suppose the test example $body \; temperature = 39\degree C, pus \; on \; tonsils = Yes$ is given. This example is covered by the meta-example $(5, Yes)$. Let us assume that the recommendation algorithm returns the recommendation as $body \; temperature \leq 2$ (eliminating the need to change the value of the attribute $pus \; on \; tonsils$). The recommendation means we should change the value of the attribute $body \; temperature$ from $39 \degree C$ to the value less than or equal to $37 \degree C$. Because the value $2$ of meta-attribute $\prescript{}{m}{(body \; temperature)}$ indicates the interval $(36.6,37]$. $\blacksquare$ \\

\section{Results}

The verification of the efficiency of the action rule and recommendation induction method consisted in:
\begin{itemize}
  \item The induction of action rules utilising of the Forward and Backward versions of the proposed algorithm. The action rule induction process was supervised by several effective \cite{wrobel2016} rule quality measures. During the experiments the number of generated rules and their characteristics were examined (Tab.\ref{t2}).
  \item Evaluation of the efficiency of actions recommended by the induced action rules. The research objective was to check how the changes of conditional attribute values, recommended by the action rules, impact the change of assigning examples from the source to the target class. In the experiment two approaches were tested. The first used only the induced action rules, the second used the meta-table and the recommendation algorithm described in Section 3.5 (Tab. \ref{t4}).
\end{itemize}
\vspace{-2mm}

The experiments were carried out on sixteen benchmarks data sets representing classification problems (Tab. \ref{t1}). In the data sets the number of decision classes was limited to two. Table \ref{t1} characterises each of the considered data set The names of the source and target classes are also given in the table. It was assumed that all conditional attributes are flexible.

All experiments were carried out in the 10-fold stratified cross-validation mode. Tables \ref{t2} - \ref{t4} contain average values calculated on all sixteen data sets. Table \ref{t5} presents the detailed characteristic of rule sets induced by SCARI (Forward version) for each considered data set. The rule induction was supervised by the C2 measure. This version of the algorithm allowed achieving the best classification and recommendation accuracy.

Table \ref{t2} contains average values of - the number of induced rules, the number of elementary conditions in rule premises, the number of elementary actions in rule premises, rule precision and rule coverage in the source and target decision classes. Moreover, the table features information about the percentage of statistically significant rules in induced rule sets. Fisher’s exact test was used to calculate the rule p-value and the False Discovery Rate \cite{benjamini1995} as the p-value correction method.

%\linenumbers

\begin{table}[ht]
    \centering
	\scalebox{0.85}
    {
		\begin{tabular}{lrrrrr}
			\hline
			data set & examples & attributes & source class & target class & \% source class \\
			\hline \hline
			car-reduced & 1594 & 6 & unacc & acc & 76 \\
			credit-a & 690 & 15 & bad & good & 56 \\
			credit-g & 1000 & 20 & bad & good & 30 \\
			diabetes-c & 768 & 8 & positive & negative & 35 \\
			echocardiogram & 131 & 11 & dead & alive & 67 \\
			heart-c & 303 & 13 & absent & present & 54 \\
			heart-statlog & 270 & 13 & present & absent & 44 \\
			hepatitis & 155 & 19 & die & live & 21 \\
			horse-colic & 368 & 22 & not-surgical & surgical & 37 \\
			hungarian & 294 & 13 & absent & present & 64 \\
			iris & 100 & 4 & setosa & versicolor & 50 \\
			monk1 & 124 & 6 & 0 & 1 & 50 \\
			mushroom & 8124 & 22 & poisonous & edible & 48 \\
			tic-tac-toe & 958 & 9 & o-wins & x-wins & 35 \\
			titanic & 2201 & 3 & deceased & survived & 68 \\
			vote & 435 & 16 & republican & democrat & 39 \\
			\hline
		\end{tabular}
    }
	\caption{Characteristic of the data sets.} \label{t1}
\end{table}

Table \ref{t4} presents the accuracy of moving examples from the source class to the target class by the action rule sets and recommendation algorithm. So far, this issue of action rule evaluation has been practically ignored in the subject literature. Below, there is a detailed description of the methodology of generating the results reported in table \ref{t2}.

The rows \textit{rule-source}, \textit{rule-target} include averages from PPV (Positive Predictive Value) calculated on test sets respectively for the source and the target class. The values were achieved using a standard rule classifier in which a given quality measure supervised the rule induction. The rows \textit{xgb-source}, \textit{xgb-target} contain average PPV values calculated on test sets for the source and the target class. The values were achieved with the use of the XGBoost \cite{chen2016xgboost} algorithm. The values are identical for each assessing measure as the algorithm does not use measures.

Table \ref{t3} shows the classification accuracy of rule-based classifiers developed based on the considered rule quality measures. The classification accuracy of the XGBoost algorithm is also presented. The rows \textit{rule class. - source}, \textit{rule class. - target} includes average values of PPV (Positive Predictive Value) calculated on test data sets, respectively, for the source and the target decision class. The values were achieved using a standard version of the rule induction algorithm (Algorithm 1). The algorithm was supervised by six ($precision$, $wLap$, $C2$, $Gain$, $Corr$, $RSS$) quality measures. The rows \textit{xgb. class. - source}, \textit{xgb. class. - target} contain average PPV values achieved with the use of the XGBoost algorithm. The values are identical for each quality measure as the algorithm does not use rule quality ones. The results illustrate the ability of rule-based classifiers as well as the XGBoost algorithm to classify test examples.

Table \ref{t4} contains information about the accuracy of changes in assigning examples from source to target classes. In other words, the results show how many examples from test sets assigned to the source class were assigned to the target class after modifications of the conditional attribute values indicated by the set of action rules. XGBoost was used as the verification algorithm. XGBoost achieves good classification results; therefore it is a reliable verifier for checking whether the attribute value modifications cause the change of assigning the example from the source to the target decision class.

The rows named $rules....$ (Tab. \ref{t4}) contain results achieved in the following way:
\begin{enumerate}
%\vspace{-2mm}
\item For a given action rule set and test example $x$ representing the source class, it was checked which of the source parts of the action rules cover $x$.
%\vspace{-2mm}
\item From the set of action rules covering $x$, the rule with the highest value of the rule quality measure of the source part was chosen.
%\vspace{-2mm}
\item The attribute values of the test example $x$ were changed according to elementary actions included in the chosen action rule premise. For elementary actions built on the basis of continuous attributes, the attribute values were changed in such a way: let us suppose the elementary action $(a \in [v_1, v_2)) \rightarrow  (a \in [v_3, v_4))$ is given, there is also given an example $x$ such that $a(x) \in [v_1, v_2)$, then the value $a(x)$ is changed to the value $(v_4-v_3)/2$. For example, if we have an elementary action $(a \in [2, 3)) \rightarrow  (a \geq 4)$ and a test example $a=2.2$, then $a:=(max_a - 4)/2$, where $max_a$ is the maximal value of $a$ in the training data set.
    %\vspace{-2mm}
\item The XGBoost algorithm classified the example achieved as such. If the example was classified to the target class, the change of attribute values was recognized as successful.
\end{enumerate}
%\vspace{-2mm}

The above strategy of changing attribute values we call \textit{the best action rule} approach.

The rows named \textit{recommendation} contain the results achieved by the recommendation algorithm described in Section 3.5. After changing the attribute values, the example was also classified with the use of the XGBoost algorithm. If the example was classified to the target class, the attribute value change was recognized as successful.

Table \ref{t7}, in turn, contains detailed information about recommendation accuracy for each data set and the Forward version of the SCARI algorithm supervised by the C2 measure.

The analysis of the achieved results starts from a quantitative analysis of the number and quality of rules. Rule number generated by successive quality measures is similar to our previous results concerning classification rule induction \cite{wrobel2016, sikora2013data}. The number of induced rules, the number of elementary conditions in rule premises and the precision and coverage of the rules depend on the applied rule quality measure.

In the Forward version of the SCARI algorithm, first the sources of elementary actions are induced. It means that for an action rule $r$, first the rule $r_S$ is induced, and next, based on the conditional attributes occurring in the rule $r_S$ premise, the rule $r_T$ is generated. The conjunctions of the sources of elementary actions comprise source parts of the action rules, while conjunctions of targets of the elementary actions –- target parts of the action rules.

Thus, it is interesting to check if the differences between the values of precision and coverage of rules $r_S$ and $r_T$ are significant. As we can see in Table \ref{t2} (rows \textit{rule precision source/target}, \textit{rule coverage source/target} - Forward method), the rule precision in the target classes is, on average, 2\% lower than the rule precision in the source classes. The rule coverage in the target classes is also lower by  2\% on average (e.g. for $precision$ measure by 3\%, and for $C2$ measure by 1\%). Thus in the Forward version of the SCARI algorithm, the quality of rules $r_S$ and $r_T$ is not significantly different.

The Backward version of the algorithm represents the reverse approach to the action rule induction. First, the target parts of the elementary actions are induced; thus, the induction is oriented towards generating high-quality target parts of the action rules (i.e. high quality $r_T$ rules). In this case (rows \textit{rule precision source/target}, \textit{rule coverage source/target} - Backward methods)) the difference between the precision of rules $r_S$ and $r_T$ can be as much as 7\% (3.6\% on average). As a consequence, for measures $precision$, $wLAp$, $C2$, $Gain$, $Corr$ the coverage of rules $r_S$ is higher than the coverage of $r_T$.

\begin{table}[ht]
	\centering
	\scalebox{0.85}{
		\begin{tabular}{p{4.2cm}lrrrrrr}
			\hline
			& method & precision & wLap & C2 & Gain & Corr & RSS \\
			\hline \hline
			rules & Forward & 14.59 & 13.88 & 11.64 & 5.82 & 5.48 & 3.62 \\
			rules & Backward & 20.57 & 19.19 & 15.07 & 6.07 & 5.05 & 3.79 \\
			conditions & Forward & 3.19 & 3.29 & 3.46 & 2.87 & 2.74 & 2.63 \\
			conditions & Backward & 3.11 & 3.23 & 3.36 & 2.83 & 2.78 & 2.55 \\
			actions & Forward & 1.58 & 1.60 & 1.45 & 0.95 & 0.91 & 0.80 \\
			actions & Backward & 2.59 & 2.70 & 2.78 & 2.37 & 2.26 & 2.00 \\
			rule precision source & Forward & 0.94 & 0.94 & 0.93 & 0.84 & 0.82 & 0.77 \\
			rule precision source & Backward & 0.88 & 0.89 & 0.89 & 0.81 & 0.81 & 0.78 \\
			rule precision target & Forward & 0.92 & 0.93 & 0.92 & 0.84 & 0.83 & 0.83 \\
			rule precision target & Backward & 0.95 & 0.95 & 0.93 & 0.85 & 0.84 & 0.80 \\
			rule coverage source & Forward & 0.28 & 0.31 & 0.43 & 0.60 & 0.63 & 0.69 \\
			rule coverage source & Backward & 0.31 & 0.36 & 0.44 & 0.62 & 0.61 & 0.61 \\
			rule coverage target & Forward & 0.26 & 0.30 & 0.42 & 0.58 & 0.59 & 0.58 \\
			rule coverage target & Backward & 0.29 & 0.32 & 0.42 & 0.58 & 0.60 & 0.66 \\
			significant FDR source & Forward & 0.83 & 0.83 & 0.78 & 0.75 & 0.73 & 0.72 \\
			significant FDR source & Backward & 0.92 & 0.96 & 0.97 & 0.98 & 0.98 & 0.97 \\
			significant FDR target & Forward & 0.77 & 0.80 & 0.74 & 0.64 & 0.65 & 0.58 \\
			significant FDR target & Backward & 0.97 & 0.98 & 0.96 & 0.96 & 0.95 & 0.96 \\
			\hline
		\end{tabular}
	}
	\caption{Characteristic of the induced sets of action rules. Average values over 16 benchmark data sets} \label{t2}
\end{table}

The Forward algorithm generates fewer elementary actions than the Backward one. In a simplified way, based on the results from 16 data sets, we may say that it is one elementary action less. The number of elementary actions decreases along with the increase of the rule coverage. In other words, the rule quality measures put emphasize on the rule coverage generate fewer elementary actions. This refers to both versions of the algorithm (Forward and Backward). The substantial majority of the induced rules are statistically significant. For example, for $precision$ rule quality measure from 78\% (Forward – target class) to 97\% (Backward – target class) rules are statistically significant.

Among the considered rule quality measures, $C2$ is the most stable one. In both versions of the algorithm (Forward, Backward), $C2$ generates rules characterised by high precision and coverage both in the source and target classes. Moreover, in the Backward version of the algorithm $C2$ generates the highest number of elementary actions.

\begin{figure}[!tpbh]%figure2
\centerline{\includegraphics[width=0.99\textwidth]{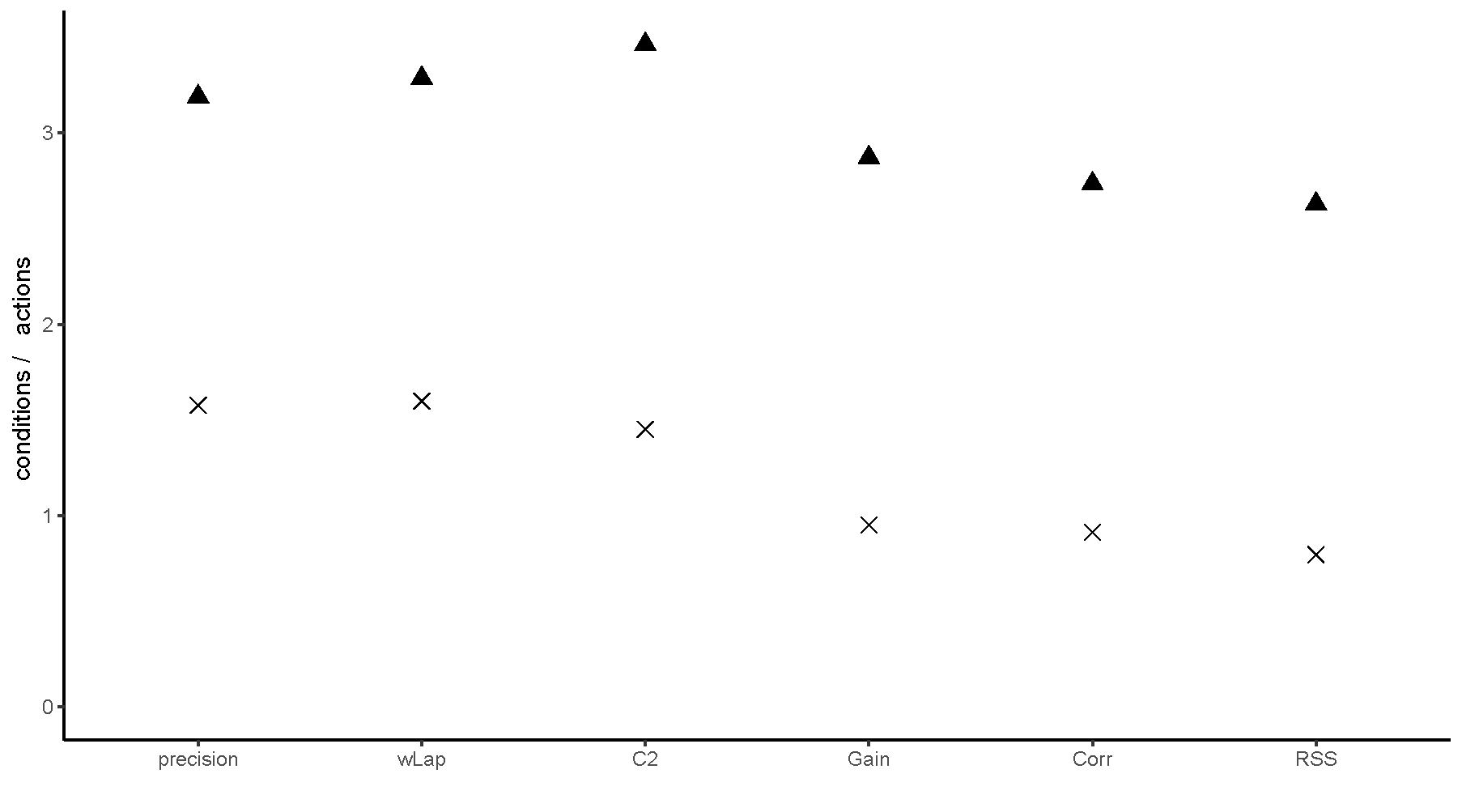}}
\caption{The average number of elementary actions (crosses) and elementary conditions (triangles) induced by the Forward version of the algorithm.} \label{conacF}
\end{figure}

\begin{figure}[!tpbh]%figure2
\centerline{\includegraphics[width=0.99\textwidth]{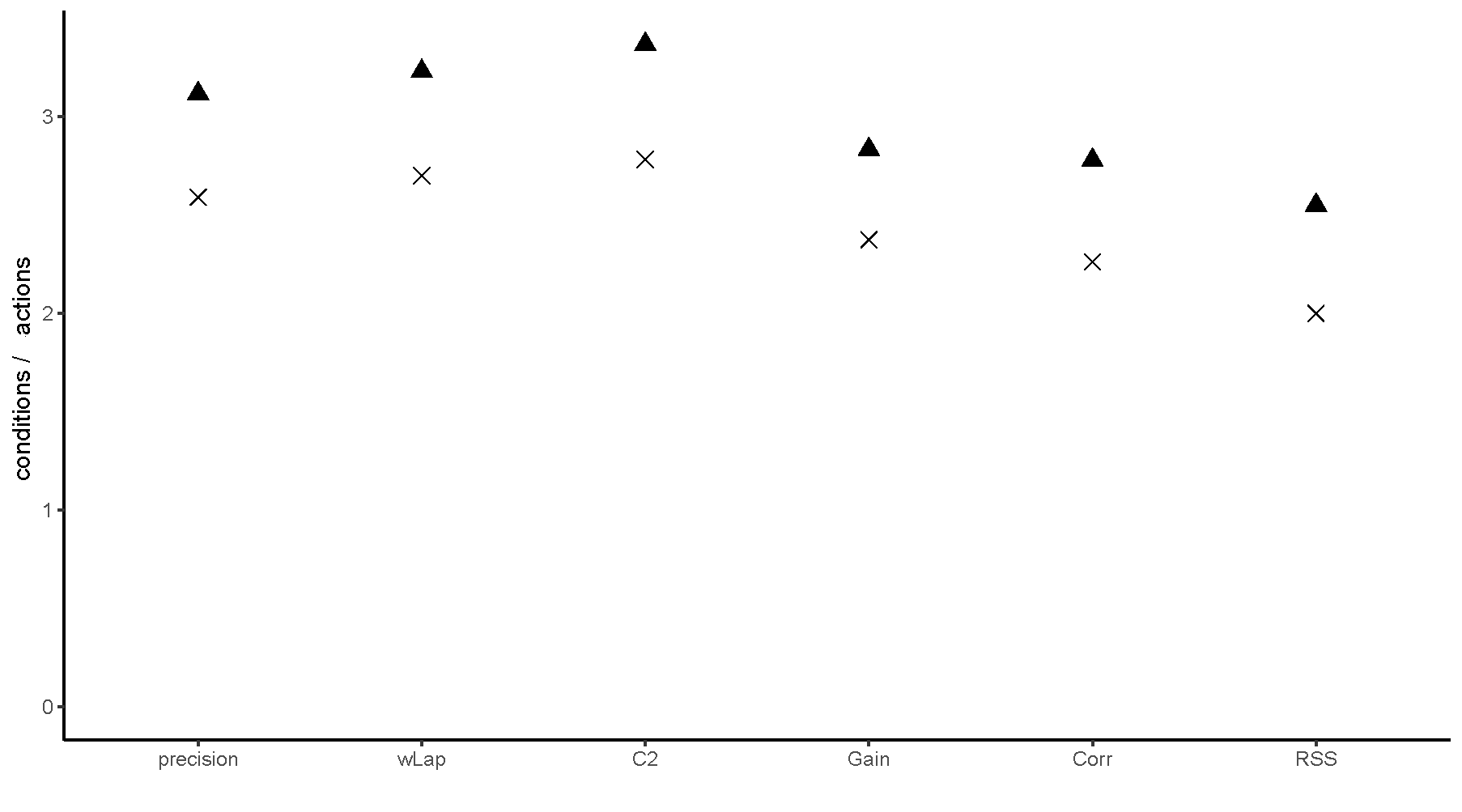}}
\caption{The average number of elementary actions (crosses) and elementary conditions (triangles) induced by the Backward version of the algorithm.} \label{conacB}
\end{figure}

\begin{figure}[!tpbh]%figure2
\centerline{\includegraphics[width=0.99\textwidth]{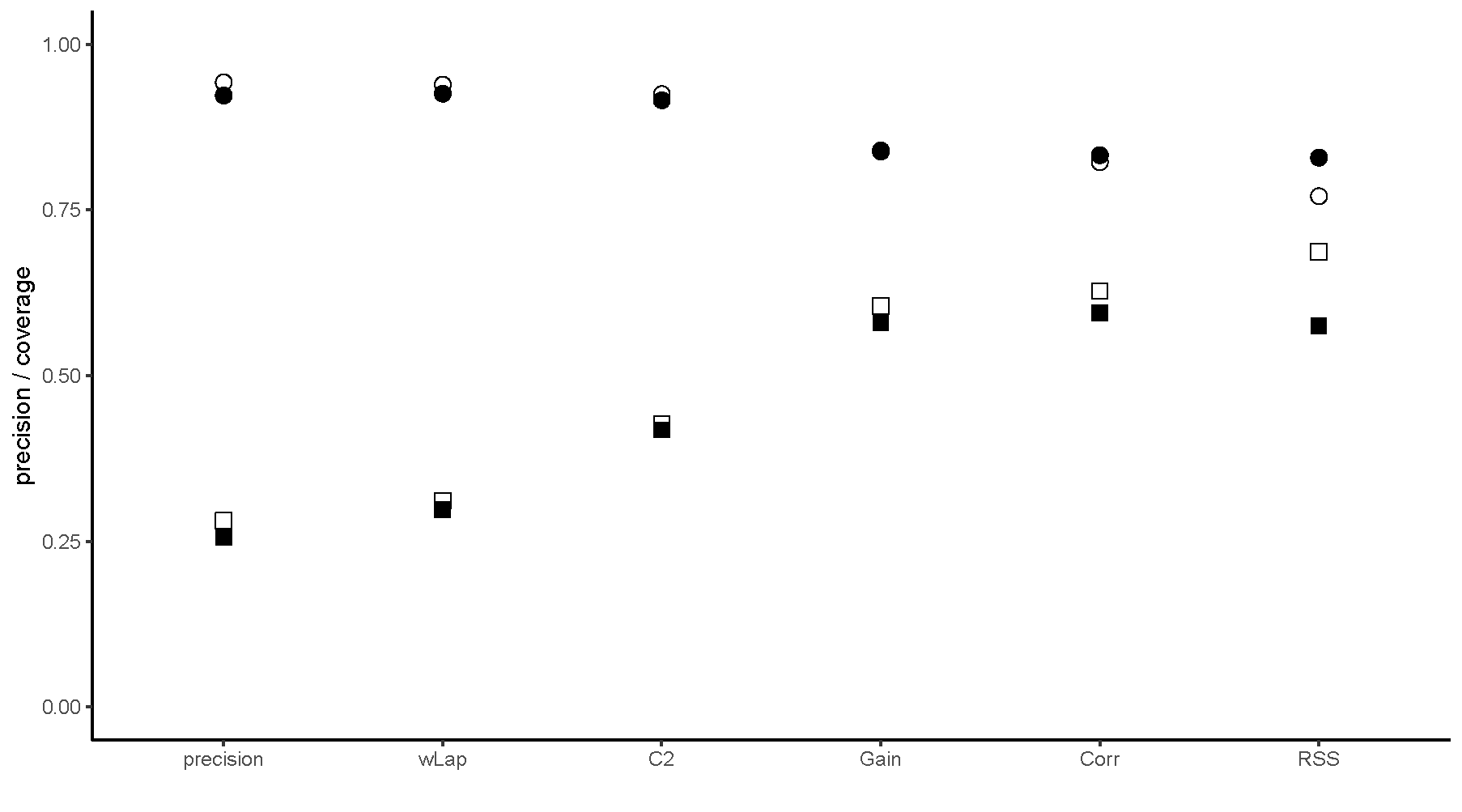}}
\caption{Average precision (circles) and coverage (squares) of the induced action rules. Forward version of the algorithm.} \label{preccovF}
\end{figure}

\begin{figure}[!tpbh]%figure2
\centerline{\includegraphics[width=0.99\textwidth]{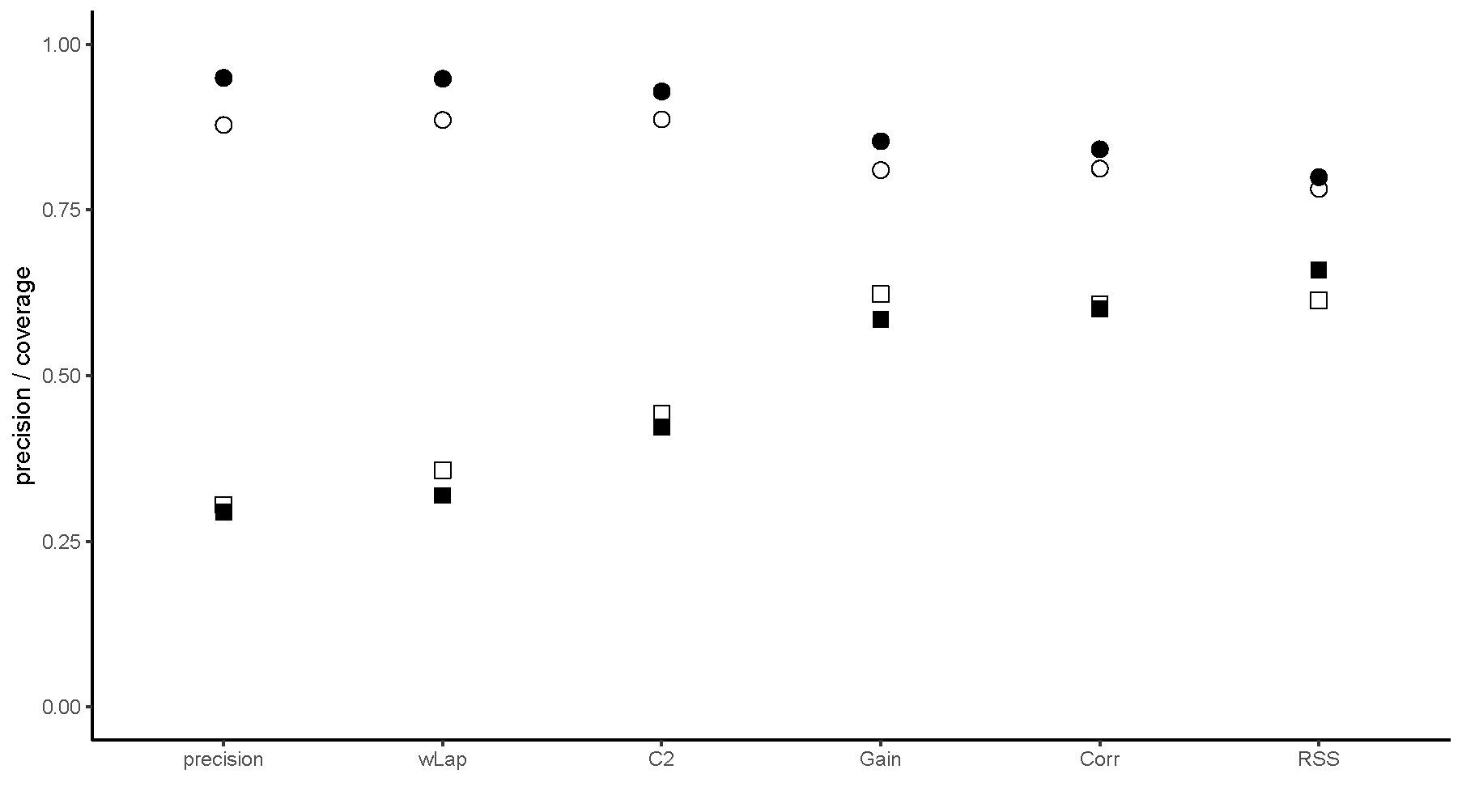}}
\caption{Average precision (circles) and coverage (squares) of the induced action rules. Backward version of the algorithm.} \label{preccovB}
\end{figure}

To better illustrate the results from Table \ref{t2}, figures \ref{conacF} and \ref{conacB} present the plots of the average number of conditions and elementary actions induced by the SCARI algorithm. Figures \ref{preccovF} and \ref{preccovB} give information about the average value of precision and coverage of generated action rules. In figures, the source class is represented by the filled circles and rectangles.

\begin{table}[ht]
	\centering
	\scalebox{0.9}{
		\begin{tabular}{lrrrrrrr}
			\hline
			classification accuracy & precision & wLap & C2 & Gain & Corr & RSS \\
			\hline \hline
			rule class. - source & 81.3 & 82.2 & 83.1 & 82.7 & 82.7 & 82.1 \\
			rule class. - target & 84.3 & 83.1 & 84.4 & 82.5 & 82.0 & 80.4 \\
            \hline
			xgb class. - source & 83.3 &  &  &  &  &  \\
			xgb class. - target & 85.1 &  &  &  &  &  \\
			\hline
		\end{tabular}
	}
	\caption{Classification accuracy of the rule-based classifiers and gradient boosting method. The results are given as a percentage.} \label{t3}
\end{table}

\begin{table}[ht]
	\centering
	\scalebox{0.9}{
		\begin{tabular}{llrrrrrr}
			\hline
			recommendation accuracy & Method & precision & wLap & C2 & Gain & Corr & RSS \\
			\hline \hline
			the best action rule  & Forward & 66.6 & 69.5 & 69.7 & 62.1 & 62.2 & 61.0 \\
			the best action rule & Backward & 50.0 & 55.0 & 63.6 & 67.9 & 63.8 & 59.9 \\
			recommendation & Forward & 60.1 & 74.3 & 82.1 & 79.5 & 78.2 & 75.7 \\
			recommendation & Backward & 62.8 & 75.7 & 85.7 & 82.8 & 81.8 & 76.8 \\
			\hline
		\end{tabular}
	}
	\caption{Recommendation accuracy. The results are given as a percentage.} \label{t4}
\end{table}

The results featured in Table \ref{t4} illustrate the efficiency (accuracy) of recommendations. One can see that good recommendation accuracy cannot be achieved if \textit{the best action rule} strategy is applied to the test examples. In this approach the XGBoost algorithm classifies to the target class slightly over 60\% out of all test examples with attribute values modified following this approach. Let us note that the XGBoost classifier classifies well the examples representing the target class (see Table \ref{t3} row \textit{xgb class. - target}.

Much better results can be achieved using the recommendation algorithm, particularly when the $C2$, $Gain$ measures and the Backward version of the algorithm are used. In this case, on average, 84.6\% and 81.9\% of examples with modified attribute values according to the recommendation algorithm suggestion are classified by XGBoost to the target class. It is interesting to note that the recommendation algorithm efficiency is the highest for the $C2$ measure. Due to the rule precision and rule coverage evaluation, $C2$ is placed in the middle of the $precsion$ measure (assessing only the rule precision) and $RSS$ (which strongly emphasizes the rule coverage).

\begin{table}[ht]
\centering
\scalebox{0.9}{
\begin{tabular}{llllllllll}
  \hline
data set & \parbox[t]{2mm}{\rotatebox[origin=c]{270}{rules}} & \parbox[t]{2mm}{\rotatebox[origin=c]{270}{conditions}} & \parbox[t]{2mm}{\rotatebox[origin=c]{270}{actions}} & \parbox[t]{2mm}{\rotatebox[origin=c]{270}{ rule precision source }} & \parbox[t]{2mm}{\rotatebox[origin=c]{270}{ rule precision target }} & \parbox[t]{2mm}{\rotatebox[origin=c]{270}{ rule coverage source }} & \parbox[t]{2mm}{\rotatebox[origin=c]{270}{ rule coverage target }} & \parbox[t]{2mm}{\rotatebox[origin=c]{270}{ \% significant source }}  & \parbox[t]{2mm}{\rotatebox[origin=c]{270}{ \% significant target }} \\
  \hline \hline
car-reduced & 9.5(0.7) & 2.0(0.1) & 1.6(0.21) & 0.97 & 0.42 & 0.17 & 0.19 & 100 & 0.99 \\
credit-a & 11.7(1.3) & 4.7(0.2) & 1.2(0.2) & 0.94 & 0.89 & 0.38 & 0.56 & 60 & 27 \\
credit-g & 42.8(0.6) & 4.6(0.3) & 1.5(0.1) & 0.87 & 0.98 & 0.06 & 0.6 & 83 & 11 \\
diabetes-c & 33.4(1.8) & 4.6(0.3) & 1.8(0.1) & 0.93 & 0.98 & 0.14 & 0.17 & 88 & 59 \\
echocardiogram & 4.0(0.8) & 2.6(0.4) & 1.0(0.0) & 0.98 & 0.96 & 0.74 & 77 & 70 & 100 \\
heart-c & 12.7(3.4) & 4.5(0.3) & 1.8(0.3) & 0.95 & 0.94 & 0.40 & 0.29 & 67 & 96 \\
heart-statlog & 11.1(1.7) & 3.6(0.4) & 1.4(0.2) & 0.96 & 0.85 & 0.34 & 0.34 & 81 & 67 \\
hepatitis & 3.8(0.8) & 4.7(0.8) & 1.5(0.3) & 0.91 & 0.98 & 0.38 & 0.45 & 98 & 49 \\
horse-colic & 8.4(1.3) & 5.0(0.3) & 1.2(0.3) & 0.81 & 0.90 & 0.52 & 0.51 & 56 & 10 \\
hun-h-disease & 9.0(1.2) & 4.3(0.3) & 1.6(0.3) & 0.94 & 0.89 & 0.48 & 0.40 & 74 & 94 \\
iris-reduced & 1.0(0.0) & 1.0(0.0) & 1.0(0.0) & 1.00 & 1.00 & 1.00 & 0.97 & 100 & 100 \\
monk1\_train & 8.0(0.7) & 2.2(0.2) & 1.1(0.2) & 0.85 & 0.96 & 0.21 & 0.42 & 28 & 91 \\
mushroom & 7.8(0.4) & 2.7(0.1) & 1.1(0.1) & 0.97 & 0.98 & 0.54 & 0.37 & 100 & 100 \\
tic-tac-toe & 16.7(1.3) & 3.5(0.2) & 2.7(0.2) & 0.90 & 0.97 & 0.12 & 0.07 & 93 & 81 \\
titanic & 4.1(0.3) & 2.5(0.0) & 1.6(0.1) & 0.82 & 0.98 & 0.52 & 0.15 & 51 & 100 \\
vote & 2.2(0.4) & 2.8(0.3) & 1.0(0.0) & 0.97 & 0.98 & 0.83 & 0.95 & 100 & 100 \\
   \hline
\end{tabular}
}
\caption{Characteristic of the induced sets of action rules. Action rule induction method -- SCARI (Backward, C2.). In parentheses the standard deviations are given.} \label{t5}
\end{table}

The recommendations induced by the recommendation algorithm are constructed on the whole set of action rules. To apply a recommendation to a given example, it is necessary to make a bigger number of attribute value changes (elementary actions) than when the best action rule approach is applied to the example.

In the case of the $C2$ measure and the recommendation induction algorithm it is required to make, on average, 2.85 elementary actions, while in the case of the best action rule approach from 1.61 (Forward action rule induction) to 2.53 (Backward action rule induction) elementary actions.

Figures 1 and 2 feature the so-called CD diagrams that compare recommendation algorithm efficiency using both versions of the SCARI algorithm and each considered quality measure. The Friedman test and the Nemenyi post-hoc test were used \cite{demvsar2006statistical} to compare multiple algorithms on multiple data sets. This comparison methodology is very conservative, and it rarely allows showing the advantage of one algorithm over another. Still, analysis the algorithm rankings shows that the $C2$ measure takes two of the first three places in the rankings. The Backward version of the algorithm achieved the best results.

Table \ref{t8} shows the results of the statistical comparison of the recommendation induction algorithm. Two versions of the algorithm were considered. In the Forward (Backward) version of the recommendation algorithm the meta-table was build on the action rule set induced by the Forward (Backward) version of the SCARI algorithm. For each rule quality measure, the comparison was made separately; therefore, the Wilcoxon signed-rank test was used in statistical analysis. One can see that for the $C2$, $Corr$, and $Gain$ measures, the difference between the Backward and Forward versions of the recommendation algorithm is statistically significant -- at the significance level of 0.1.

The presented results show that the C2 measure and Backward approach to the action rule induction allows generating sets consisting of a moderate number of action rules. These rule sets contain rules of good precision and coverage - in both source and target classes. The recommendation algorithm basing on the sets of these rules permit good recommendation accuracy. Table \ref{t7} presents the recommendation accuracy achieved for each of the considered 16 data sets.

\begin{table}[ht]
	\centering
	\scalebox{0.99}{
		\begin{tabular}{p{4.2cm}lccc}
			\hline
			& SCARI & SCARI & ARED \\
		    & Forward & Backward & ARED \\
			\hline \hline
			rules & 15.65 & 20.72 & 3308.13 \\
			conditions & 3.84 & 3.81 & 2.38 \\
			actions & 1.61 & 2.53 & 2.38 \\
			rule precision source & 0.88 & 0.88 & 0.54 \\
			rule precision target & 0.91 & 0.90 & 0.69 \\
			rule coverage source & 0.30 & 0.29 & 0.19 \\
			rule coverage target & 0.25 & 0.30 & 0.27 \\
			\hline
		\end{tabular}
	}
	\caption{Characteristic of the action rule sets induced by the SCARI and ARED algorithms. In the SCARI algorithm, rule precision supervised the rule induction process.} \label{t6}
\end{table}

The last of the presented tables shows a quantitative comparison between the SCARI and the ARED algorithms. ARED requires discretised data; therefore before action rule induction, all data sets were discretised. In the experiments we used our ARED algorithm implementation, as we could not find any available implementation on the Internet. Our implementation of ARED is available in the GitHub repository.

\begin{table}[ht]
\centering
\begin{tabular}{llc}
  \hline
data set & rules & recommendation \\
  \hline \hline
car-reduced & 74.6 & 99.4 \\
credit-a & 45.8 & 76.5 \\
credit-g & 77.7 & 67.3 \\
diabetes-c & 69.6 & 83.0 \\
echocardiogram & 78.7 & 97.2 \\
heart-c & 57.0 & 47.3 \\
heart-statlog & 78.8 & 58.0 \\
hepatitis & 83.3 & 80.0 \\
horse-colic & 57.0 & 92.6 \\
hun-h-disease & 55.0 & 81.3 \\
iris-reduced & 95.0 & 100 \\
monk1 & 39.2 & 100 \\
mushroom & 25.6 & 95.3 \\
tic-tac-toe & 56.7 & 94.7 \\
titanic & 45.1 & 100 \\
vote & 80.3 & 98.8 \\
   \hline
\end{tabular}
\caption{Recommendation accuracy. Action rule induction method -- SCARI (Backward, C2). The results are given as a percentage.} \label{t7}
\end{table}

\begin{figure}[!tpbh]%figure2
\centerline{\includegraphics[width=0.99\textwidth]{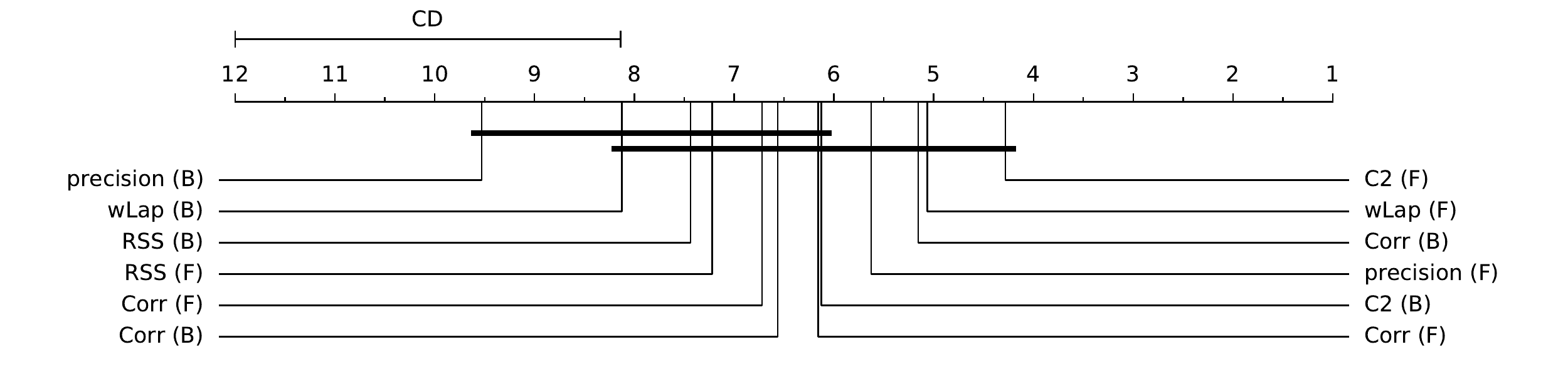}}
\caption{Recommendation accuracy. Critical difference diagram for action rule based recommendations.}\label{cd-rules}
\end{figure}

\begin{figure}[!tpbh]%figure2
\centerline{\includegraphics[width=0.99\textwidth]{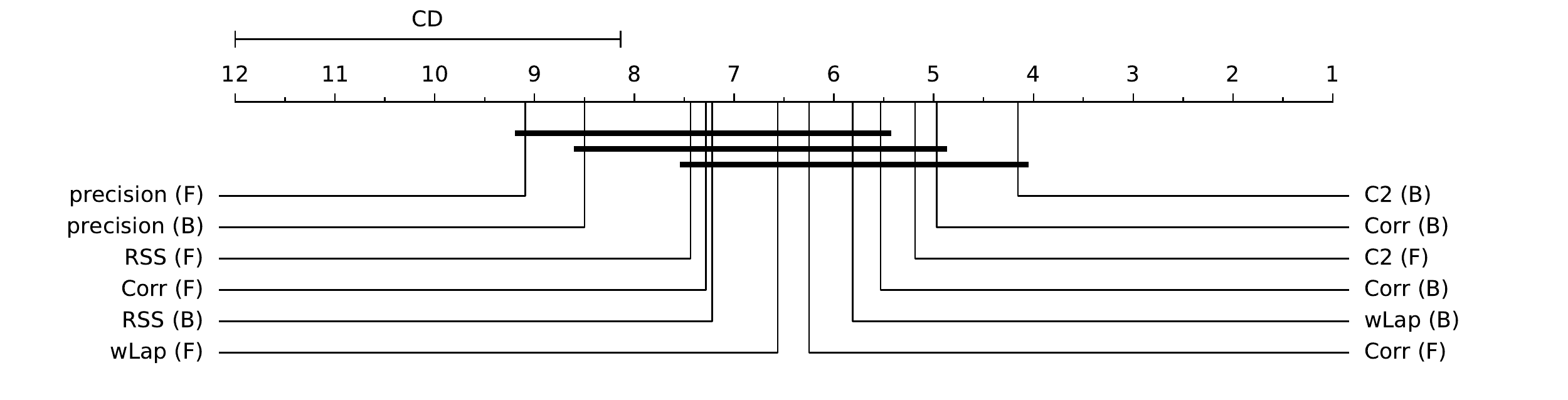}}
\caption{Recommendation accuracy. Critical difference diagram for recommendation induction algorithm.}\label{cd-recom}
\end{figure}

\begin{table}[ht]
\centering
\begin{tabular}{lcccccc}
  \hline
quality & Forward & Forward & Backward & Backward & Wilcox & Wilcox \\
measure & (avg) & (median) & (avg) & (median) & (p-value)& (FDR) \\
  \hline \hline
precision & 60 & 56 & 63 & 60 & 0.398 & 0.597 \\
wLap & 74 & 86 & 76 & 84 & 0.756 & 0.756 \\
C2 & 82 & 94 & 66 & 94 & 0.032 & 0.088 \\
Gain & 80 & 82 & 83 & 84 & 0.044 & 0.088 \\
Corr & 78 & 83 & 82 & 83 & 0.024 & 0.088 \\
RSS & 76 & 82 & 77 & 81 & 0.610 & 0.732 \\
   \hline
\end{tabular}
\caption{Statistical comparison of the Forward and Backward versions of the SCARI algorithm. Each measure is considered separately. The results are given as a percentage.} \label{t8}
\end{table}

The rest of this section contains examples of recommendations induced for three real-life examples. All examples refer to benchmark data sets. The sets characteristics and detailed explanations of the meaning of the attributes can be found in the UCI MLDB repository. The seismic bumps data set, in particular, was made available by the authors of this article a few years ago. The recommendation generated for an example from this set will be discussed in more detail at the end of this section.In the experiment, all attributes were considered as flexible.

The first example refers to the Diabetes-c data set. The meaning of conditional attributes is the following: $Preg$ -- number of times pregnant, $Plas$ -- plasma glucose concentration a 2 hours in an oral glucose tolerance test, $Pres$ -- diastolic blood pressure, $Skin$ -- Triceps skinfold thickness, $Insu$ -- hour serum insulin, $Mass$ -- body mass index, $Pedi$ -- diabetes pedigree function, $Age$ -- age. The source class indicates examples \textit{tested positive for diabetes} while the target class indicates examples \textit{tested negative}.

During the experiment a test example with the following attribute values was considered: $Preg=8$, $Plas=183$, $Pres=64$, $Skin=0$, $Insu=0$, $Mass=23.3$, $Pedi=0.672$, $Age=32$. This example is covered in meta-table by the following meta-example: $Preg > 7$, $Plas \geq 154$, $Pres \in [57, 84.5]$, $Skin < 31$, $Insu < 193$, $Mass \leq 32$, $Pedi < 1.2$, $Age > 29$. The highest-quality recommendation inducted for the example contains only one elementary action: $Plas > 154 \rightarrow Plas < 123$.

According to the elementary action, the value $Plas$ was changed to 106. The number 106 lies in the middle of the interval $[mPlas, 123)$, where $mPlas$ is the minimal value of the attribute $Plas$. The remaining attribute values did not change. The XGBoost algorithm classified such a changed example to the target class. It is also worth to note that the target part od the elementary action $Plas > 154 \rightarrow Plas < 13$ covers 281 examples from the target class and 64 examples from the source class.

The second example refers to the Congressional voting data set \cite{schlimmer1987concept}. All attributes in the set are binary. Therefore the meta-table is identical to the original data table (the original data set). The set contains the following attributes: \textit{handicapped infants, water project cost sharing, adoption of the budget resolution, physician fee freeze, el salvador aid, religious groups in schools, anti satellite test ban, aid to nicaraguan contras, mx missile, immigration, hsynfuels corporation cutback, education spending, superfund right to sue, crime, duty free exports, export administration act south africa}. The source class indicates examples labeled as \textit{republican} while the target class indicates examples labeled as \textit{democrat}.

The following example, representing the $republican$ decision class, was analyzed:
\textit{handicapped-infants=n, water-project-cost-sharing=y, adoption-of-the-budget-resolution=n, physician-fee-freeze=y, el-salvador-aid=y, religious-groups-in-schools=y, anti-satellite-test-ban=n, aid-to-nicaraguan-contras=n, mx-missile}

\noindent \textit{=n, immigration=n, synfuels-corporation-cutback=n, education-spending=y,}

\noindent \textit{superfund-right-to-sue=y, crime=y, duty-free-exports=n, export-administration-act-south-africa=n}.

The highest-quality recommendation induced for this example contains two elementary actions:
($crime=y \rightarrow crime=n$) \textbf{and} ($adoption-of-the-budget-resolution=n \rightarrow adoption-of-the-budget-resolution=y$).

The example with attribute values changed according to the above recommendation was classified by the XGBoost algorithm to the target decision class.
It is also worth noting that the conjunction $crime=n$ \textbf{and} $adoption-of-the-budget-resolution=y$ covers 156 examples labelled as \textit{democrat} and three examples representing the \textit{republican} decision class.

The last example concerns the assessment of seismic hazards in mines. The data set describes 8-hour work cycles of a mine \cite{sikora2010application}. The aggregated values from the seismic and acoustic measurement systems are the basis of the seismic hazard assessment. Two classes were distinguished in the data set: the target class reflecting a normal state (\textit{no hazard}) and the source class reflecting a \textit{hazardous} state.

The meaning of particular attributes is as follows: $Seismic$ -- result of shift seismic hazard assessment in the mine working obtained by the so-called seismic method ($a$ -- lack of hazard, $b$ -- low hazard, $c$ -- high hazard, $d$ -- danger state); $Seismoacoustic$ -- result of shift seismic hazard assessment in the mine working obtained by the acoustic method; $Shift$ -- information about the type of a shift ($W$ -- coal-getting, $N$ -preparation shift); $GEnergy$ -- seismic energy recorded within a previous shift by the most active geophone ($GMax$) out of geophones monitoring the longwall; $GPuls$ -- a number of pulses recorded within a previous shift by $GMax$; $GDEnergy$ -- a deviation of energy recorded within a previous shift by $GMax$ from average energy recorded during eight previous shifts; $GDPuls$ -- a deviation of a number of pulses recorded within a previous shift by $GMax$ from average number of pulses recorded during eight previous shifts; $GHazard$ -- result of shift seismic hazard assessment in the mine working obtained by the seismoacoustic method based on registration coming from $GMax$ only; $NBumps$ -- the number of seismic bumps recorded within a previous shift; $NBumps2$ -- the number of seismic bumps (in energy range $[10^2,10^3)$) registered within a previous shift ($NBumps2$,...,$NBums9$ –- have the analogical meaning as the nbumps2 attribute); $Energy$ -- total energy of seismic bumps registered within a previous shift; $MEnergy$ -- the maximum energy of the seismic bumps registered within a previous shift.

The following example, representing the \textit{hazardous} state, was analysed: $seismic=a$, $seismoacoustic=b$, $shift=N$, $GEnergy=92520$, $GPuls=169$, $GDEnergy=-73$, $GDPuls=-74$, $GHazard=a$, $NBumps=1$, $NBumps2=0$, $NBumps3=0$, $NBumps4=1$, $NBumps5$,...,$NBumps9=0$, $Energy=10000$, $MEnergy=10000$.

This example is covered in meta-table by the following meta-example:

\noindent $seismic=a$, $seismoacoustic=b$, $shift=N$, $GEnergy \in [60115, 376395)$, $GPuls \in [55, 334)$, $GDEnergy  \in [-29, 88)$, $GDPuls < -42$, $GHazard=a$, $NBumps \in [1, 4]$, $NBumps2 \in [0, 2]$, $NBumps3 \leq 2$, $NBumps4 \leq 1$, $NBumps5$,...,$NBumps9=0$, $Energy > 6050$, $MEnergy > 5500$.

The highest-quality recommendation induced for this example contains one elementary action: $GPuls \in [54, 195] \rightarrow GPuls < 32$.
The target part of this elementary action covers 153 examples representing the \textit{no hazard} decision class and no examples form the \textit{hazardous} class.

This time, for the considered example, the value of the $GPules$ attribute was changed slightly below 32. This is the smallest change required to fulfil the recommendation. After the change, the example was classified to the \textit{no hazard} class by the XGBoost algorithm.

The change suggested by the recommendation means that in order to lower the hazard state, it is necessary to decrease the number of impulses registered by geophones. In practice, it means that the mining process is slower, which, in turn, may result in stress relief of the rock mass. The rock mass stress relief results in lowering the hazard. Thus the recommendation is justified in terms of domain knowledge. In this example we did not take into consideration the economic consequences of slowing down the mining process.

\section{Conclusions}
The article presents the SCARI algorithm, which allows action rules induction using the sequential covering strategy. The algorithm can use rule quality measures to supervise the rule induction process. Two versions of the algorithm were proposed: Forward and Backward. The generated rule sets represent dependencies found in data and are treated as a result of the exploratory data analyses. Moreover, the induced set of action rules are the basis for two strategies of changing attribute values: the best quality rule and the recommendation induction algorithm.

The recommendation induction algorithm is an adapted version of the standard rule induction algorithm. The algorithm works on meta-examples -- elements of the meta-table. The meta-table is generated based on cuts defined by the elementary conditions of elementary actions, a kind of analogy to the discretisation of numerical attributes. Let us note that after discretisation of numerical attributes, the ordinal attributes occur in a discretised data set. In such a data set, it is still sensible to generate a meta-table for a recommendation algorithm.

The proposed methods were experimentally verified on the basis of 16 benchmark data sets. Quantitative and qualitative analysis of the induced action rule sets was carried out. The Backward version of the SCARI algorithm, supervised by the C2 measure, proved very effective. The XGBoost algorithm was used to verify the recommendation accuracy. The experiments carried out showed that examples with changed attribute values indicated by the recommendation algorithm are mostly classified as the examples representing the target class. At the end of the section describing the experimental results, three real-life examples of recommendations were presented.

The recommendation induction algorithm can generate many recommendations of different quality. In the conducted research, the best recommendation was always applied to the test example. The best recommendation “moves” an example into the area of the feature space covered by many examples representing the target class and few representing the source class. Depending on the applied quality measure, recommendations can be more specific or more general. In our future work we plan to introduce the cost criterion for changing the attribute values into the recommendation induction algorithm.

Our future works will focus on applying the proposed methodology to complex AI/ML models (e.g., ensemble models). Our idea follows from the assumption that a rule-based model will approximate the decisions of a complex model. The purpose of the rule-based approximator is to approximate with the predefined precision the decisions made by the complex model. Thus the approximator does not need to have good generalization abilities and can be over-fitted to decisions taken by the complex model. In practice, that the action rule induction algorithm should generate rules whose source and target parts cover only positive examples. This can be achieved by using the rule precision measure during the rule induction. The recommendations generated based on an induced set of action rules will indicate changes of attribute values sufficient for the complex model to classify an example with changed attribute values to the target class.

The source code and executable versions of the proposed algorithms are available in the GitHub repository (https://github.com/adaa-polsl/SCARI). The SCARI algorithm is part of the RuleKit package \cite{gudys2020rulekit}, a comprehensive suite for rule-based learning.

\section{Acknowledgements}
This work was partially supported by Polish National Centre for Research and Development within the Operational Programme Intelligent Development (POIR.01.01.01-00-0871/17) and Computer Networks and Systems Department at Silesian University of Technology within the statutory research project.

%\section*{References}

\bibliography{references}

\end{document}